\documentclass[conference]{IEEEtran}
\usepackage{amsmath,amsfonts}
\usepackage{algorithmic}
\usepackage{algorithm}
\usepackage{array}
\usepackage{textcomp}
\usepackage{stfloats}
\usepackage{url}
\usepackage{verbatim}
\usepackage{graphicx}
\usepackage{cite}
\usepackage{xcolor}
\usepackage{booktabs}
\usepackage{graphicx}
\usepackage{subfigure}
\usepackage{tikz}
\usepackage{multirow}

\begin{document}

\title{LOID: Lane Occlusion Inpainting and Detection for Enhanced Autonomous Driving Systems}

\author{
    \IEEEauthorblockN{Aayush Agrawal\IEEEauthorrefmark{1}, 
                      Ashmitha Jaysi Sivakumar\IEEEauthorrefmark{1}, 
                      Ibrahim Kaif\IEEEauthorrefmark{1}, 
                      Chayan Banerjee\IEEEauthorrefmark{2}}
    \IEEEauthorblockA{\IEEEauthorrefmark{1}Team Abhiyaan, Center For Innovation, Indian Institute of Technology Madras,  India}
    \IEEEauthorblockA{\IEEEauthorrefmark{2}School of Electrical Engineering and Robotics, Queensland University of Technology, Brisbane, Australia}
      % \{agrawal, ce21b024, be21b019\}@smail.iitm.ac.in \& {c.banerjee@qut.edu.au}
}

\markboth{Journal of \LaTeX\ Class Files,~Vol.~14, No.~8, August~2021}%
{Aayush Agrawal \MakeLowercase{\textit{et al.}}: LOID: Lane Occlusion Inpainting and Detection for Enhanced Autonomous Driving Systems}
% {Aayush Agrawal \MakeLowercase{\textit{et al.}}: Enhanced Lane Detection under Occlusions: An Inpainting Approach with YOLOP}

\IEEEpubid{0000--0000/00\$00.00~\copyright~2021 IEEE}

\maketitle

\footnotetext[1]{agrawal@smail.iitm.ac.in, ce21b024@smail.iitm.ac.in, be21b019@smail.iitm.ac.in, \& c.banerjee@qut.edu.au}

\begin{abstract}
Accurate lane detection is essential for effective path planning and lane following in autonomous driving, especially in scenarios with significant occlusion from vehicles and pedestrians. Existing models often struggle under such conditions, leading to unreliable navigation and safety risks. We propose two innovative approaches to enhance lane detection in these challenging environments, each showing notable improvements over current methods.
The first approach \textit{aug-Segment} improves conventional lane detection models by augmenting the training dataset of CULanes with simulated occlusions and training a segmentation model. This method achieves a 12\% improvement over a number of SOTA models on the CULanes dataset, demonstrating that enriched training data can better handle occlusions, however, since this model lacked robustness to certain settings, our main contribution is the second approach, \textit{LOID} Lane Occlusion Inpainting and Detection. LOID introduces an advanced lane detection network that uses an image processing pipeline to identify and mask occlusions. It then employs inpainting models to reconstruct the road environment in the occluded areas. The enhanced image is processed by a lane detection algorithm, resulting in a 20\% \& 24\% improvement over several SOTA models on the BDDK100 and CULanes datasets respectively, highlighting the effectiveness of this novel technique.
\end{abstract}

% \keywords{Lane Detection, Heavy Occlusion, Image Inpainting, Autonomous Driving, Computer Vision, Deep Learning}

\begin{figure*}[ht]
    \centering
    \includegraphics[width=\linewidth]{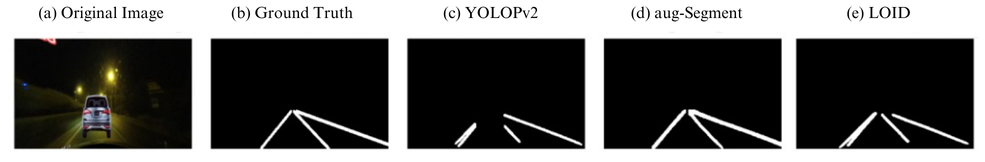} 
    \caption{Semantic Mask Generated by Different Models.}
     \label{fig:example_image}
\end{figure*}

\section{Introduction}
% \textcolor{cyan} {What is lane detection}
% \textcolor{cyan} {Importance of lane detection in autonomous driving}
% \textcolor{cyan}{Benefits for intelligent cars (adaptive cruise control, lane departure warning, traffic understanding)}

Lane detection, as a computer vision problem, involves identifying the boundaries of lanes on a roadway using visual data, typically captured by cameras mounted on a vehicle \cite{haque2019computer}. The core objective of lane detection is to accurately recognize lane markings, which can be used to determine the vehicle’s position within the lane and guide its trajectory \cite{6338738}. It offers significant benefits for intelligent vehicles, in terms of adaptive cruise control \cite{Tian2021}, lane keeping \cite{8442051,rao2022developing} and departure warning systems \cite{8305846}. This process is also essential for the navigation and control of autonomous vehicles \cite{6822610,islam2019vision}. 

However, traditional lane detection models often struggle in environments where lanes are heavily obscured by vehicles and pedestrians, which is referred to as occlusion. These occlusions disrupt the continuity of lane markings visible to the sensors in complex urban environments or dense traffic, leading to unreliable detection and potentially hazardous driving decisions. 

Therefore, addressing the challenges posed by occlusions is paramount for enhancing the robustness and reliability of autonomous driving systems. Effective lane detection in the presence of occlusions requires innovative approaches that can intelligently handle partial or obscured lane markings, ensuring continuous and precise localization of lane boundaries \cite{xiong2020fast,cao2019lane}.

Traditional lane detection techniques are categorized into model-based \cite{chen2010block,mammeri2014lane} and learning-based methods \cite{li2019line,qin2020ultra,pan2018spatial}. Model-based methods typically utilize mathematical models like edge detection and the Hough transform but struggle with varying road conditions and lighting changes. Learning-based methods, which include deep learning models, semantic segmentation, anchor-based, and parameter prediction-based, offer robust lane delineation but face challenges with occlusions and unforeseen/ novel road conditions. We discuss these topics in greater detail and present an overview of the state-of-the-art in the following section.

It is seen that traditional approaches often lack the adaptability required to handle unexpected scenarios, such as non-standard lane markings or irregular road conditions, particularly in situations where portions of lanes are obscured or fragmented due to occluding objects. Although deep learning-based methods show promise in addressing these challenges, they introduce other complexities, such as the need for large annotated datasets, the difficulty of model interpretability, and variation between road conditions in train and application data. 
As a result, while advances are being made, the mitigation of occlusions remains an ongoing challenge in lane detection research \cite{HAO2023135}.

We proposed two approaches to mitigate this problem.
Our first approach named ``aug-Segmnet’’ involves training an off-the-shelf segmentation model with a custom dataset with lane markings over occlusions. This custom-augmented dataset was developed to aid the model in handling occlusions more effectively. %While aug-Segmnet, trained on this dataset, showed some improvement over state-of-the-art models, the performance boost was relatively modest.
In our second approach called  LOID (Lane Occlusion Inpainting and Detection), we addressed the problem by introducing a novel pipeline that begins with detecting occlusions, followed by inpainting to restore the lane information. Finally, a segmentation model masks the lane lines ensuring robust lane detection. The segmentation outputs from aug-Segment and LOID in comparison to YOLOPv2 can be seen in Fig. \ref{fig:example_image}.  

Our primary contributions are listed as follows:
\begin{itemize}
    \item We leverage a custom dataset, which is augmented with occlusions, to train a YOLOv8-seg \cite{yolov8_ultralytics} model (aug-Segment), which exhibits a performance improvement on multiple SOTA models.
    \item We introduce a deep learning-based pipeline, named LOID (Lane Occlusion Inpainting and Detection), which consists of three components: detection of dynamic occlusions, reconstruction of obscured lane markings to provide a complete and coherent representation of the road, and subsequent detection of clear lane lines. LOID outperforms multiple SOTA models and aug-Segment by a substantial margin. LOID also shows good adaptability and robustness to several popular datasets.
\end{itemize}

The rest of the paper is organized as follows, section \ref{sec:related_work} analyses previous research for lane detection, section \ref{sec:method} outlines the developed pipeline and an analysis of each node in more detail, section \ref{sec:data_eval} describes the datasets used for evaluation of the study, section \ref{sec:experiments} has all the experiments and results conducted to assess the performance of the model compared to popular models.

\begin{figure*}
    \centering
    \includegraphics[width = \linewidth, height=0.3\linewidth]{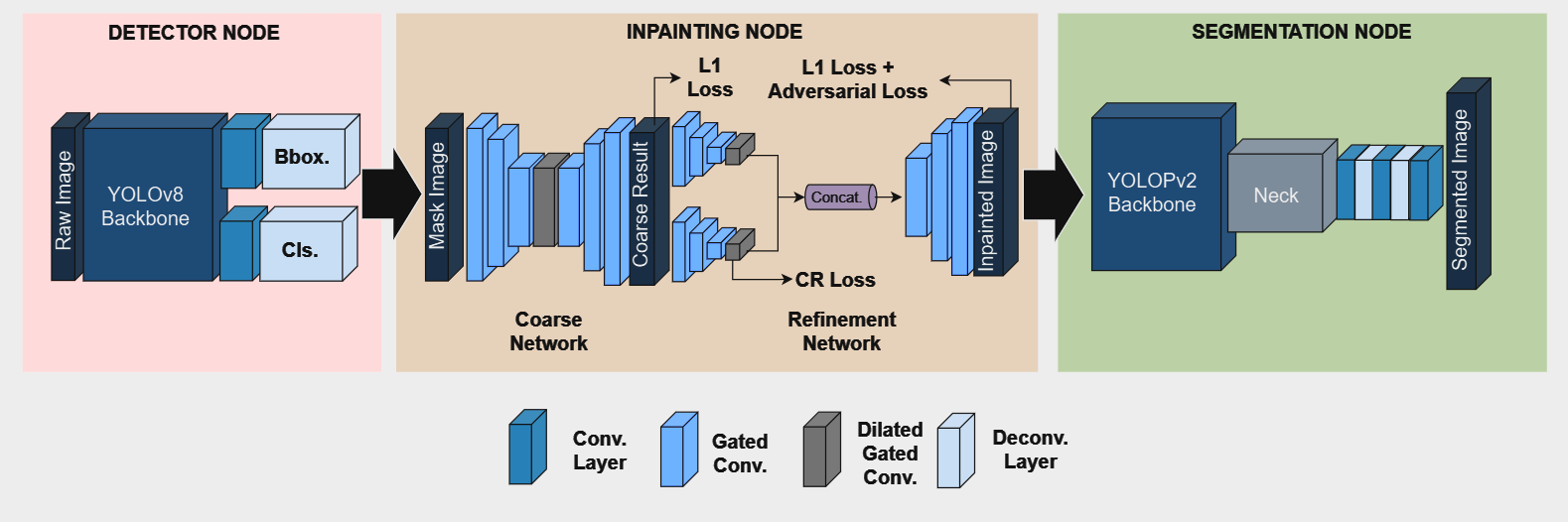}
    \caption{Overview of LOID Architecture. The detector node is responsible for identifying and classifying the occlusions. The inpainting node uses a Coarse and Refinement network to regenerate occlusion-free markings. Finally, a segmentation node is used to generate the final lane masks.}
    \label{fig:model_arch}
\end{figure*}

\section{Related Work}
\label{sec:related_work}

Conventional approaches to lane detection can be broadly categorized into two main techniques: model-based methods and learning-based methods.

% Requires more refs
\textbf{Model-Based Methods:} Model-based lane detection methods rely on mathematical models and heuristics to identify lane markings in images or video frames. These techniques often utilize image processing operations such as edge detection, the Hough transform, and curve fitting to extract lane features \cite{chen2010block,mammeri2014lane}. While model-based methods can provide rapid computation speeds, they can struggle with complex or varying road conditions that violate the assumptions of the underlying mathematical models. Such methods lack generalization and struggle in cases of lighting variations.

Methods proposed in \cite{ding2017fast} face a dependency on the camera's parameters and initial calibration. Hence, the creation of the bird's eye view of a frame is difficult to generalize for different systems and often works under the assumption of flat roads. To combat this, a deep-learning model that generated the bird's eye view and its features using a Spatial Transformation Pyramid module was proposed in \cite{wang2023bev}. Another solution was presented in \cite{luo2023latr} to reduce depth ambiguity in monocular images by directly utilizing 3D-aware front-view features and dynamic 3D ground positional embeddings, without relying on explicit camera parameters or transformed view representations.

\textbf{Learning-Based Methods:} 
In contrast, learning-based lane detection methods employ deep learning techniques to learn the visual patterns of lane markings from training data \cite{li2019line, qin2020ultra, pan2018spatial}. Deep learning models are broadly split into 4 groups- segmentation-based, anchor-based, row-wise detection, and parameter prediction methods. Semantic segmentation-based methods are either semantic \cite{8901686, han2022yolopv2betterfasterstronger} or instance \cite{neven2018endtoendlanedetectioninstance} based and are most commonly used. These methods can provide pixel-level classification for entire scenes, ensuring detailed lane delineation and robust performance. 

For Anchor-based, these methods have been widely explored in \cite{qin2022ultra, 9795098} by introducing a row anchor-based method that partitions the image into a grid and determines the presence of lane lines in each row. In row-wise detection \cite{Yoo_2020_CVPR_Workshops}, the model divides an image into rows to obtain a row-wise representation for each lane. Parameter prediction methods try to estimate lanes using polynomial curves. PolyLaneNet \cite{tabelini2021polylanenet} regarded lane detection as a curve equation and used a deep network to regress the lane curve equation, \cite{liu2021end} used the transformer structure to predict the parameters in the lane. However, real-life lanes are not always representable by curves such as diverging lanes, sidewalks, etc. \cite{ingale2016comparative}
CNNs \cite{pizzati2019lanedetectionclassificationusing}, RNNs \cite{8883072}, transformers \cite{10378626}. Self-Supervised Learning \cite{nie2024lanecorrectselfsupervisedlanedetection}, Anchor-Free Methods \cite{efrat20203dlanenetanchorfreelane}, Attention Mechanisms \cite{10248876} and  3D Lane Detection \cite{9008811} are other explored learning-based methodologies in the domain of lane detection.

% Refine this paragraph with more technical details
The challenge with these methods arises in scenarios of heavy occlusions and novel road conditions (unforeseen during training time). 
% Deep learning models, which infer lane distributions from the entire image, often struggle when lanes are partially visible or are occluded. 
This leads to a significant performance drop since occlusions can cause large portions of the road to remain unmarked when predicting lane markings. This phenomenon has been quantified in further experiments section of this paper.

Occlusions were tackled in \cite{xiong2020fast} using a multi-camera fusion framework with calibrated front, left and right-view cameras through which model-based lane detection was performed. But it required a significant amount of camera parameters, position information, and tuning. Thus, lacking robustness and portability. \cite{cao2019lane} utilized a mask operation to filter out interfering pixels and sharpen the target lane in the image and employed a third-order B-spline curve model for lane-fitting, accommodating the curvature and variability of real-world roads. To ensure robustness, the RANSAC algorithm was utilized for iterative curve fitting, which reduced the influence of outliers. But these models faced immense challenges when the road was at an elevation or downslope.

Inpainting models, forms an essential part of our proposed pipeline. 
Inpainting models typically reconstruct masked regions in an image by analyzing the surrounding pixels. These models are categorized into sequential, CNN, and GAN-based approaches, see \cite{elharrouss2020image} for more on automatic image inpainting. In a number of works inpainting models have been used for occlusion removal. For example, inpainting models were utilized in \cite{zhang2023optimized} to reconstruct masked regions of dynamic objects to better represent the static objects. \cite{8963753} presents another application, where inpainting models were utilized to refine semantic segmentation of the background by masking foreground objects and reconstructing the scene.

%While picking the inpainting node our major focus was on choosing a model that was lightweight and allowed tuning. 

%In \cite{elharrouss2020image} automatic image inpainting models were reviewed categorizing methods into sequential-based, CNN-based, and GAN-based approaches. It provides a list of techniques for different image distortions and available datasets. The paper also includes performance evaluations of these methods using standard metrics. 

\section{Methodology}
\label{sec:method}
In this section we present a comprehensive overview of aug-Segment and LOID, detailing the modifications and tuning adjustments made for performance improvement.

\subsection{Architecture Design -- aug-Segment}
An off-the-shelf instance segmentation model, YOLOv8-seg \cite{yolov8_ultralytics} pre-trained on the COCO dataset \cite{lin2014microsoft}, was fine-tuned utilizing a custom augmented dataset explained in Section \ref{cul_train}. Since the dataset is characterized by labeling present even in regions obscured by occlusions, the model learns to delineate the lanes through these occlusions.

The model was trained for 50 epochs, with an SGD optimizer set at 0.01 learning rate. The loss function for YOLOv8-seg integrates 4 components into it - Binary Cross-Entropy (BCE) Loss for class prediction, Distribution Focal Loss (DFL) for enhancing the distribution of predictions around the target to improve bounding box precision, Complete Intersection over Union (CIoU) Loss for refining bounding box predictions by considering overlap, distance, and aspect ratio differences between predicted and ground truth boxes.

\subsection{Architecture Design -- LOID}
To combat the problem of occlusions causing a loss of information, the proposed pipeline, as shown in Fig. \ref{fig:model_arch}, was constructed with three key nodes: Detection, Inpainting, and Segmentation. 

\begin{itemize}
    \item \textbf{Detector Node:} The pipeline begins with the detector node, which identifies and localizes occlusions on the road, subsequently, generating its masks. This process employs a YOLOv8 \cite{yolov8_ultralytics} object detector model.  
    
    \item \textbf{Inpainting Node:} The next part of the pipeline is the inpainting node. It receives the masked images from the detector node and passes them through CR-Fill \cite{zeng2021crfillgenerativeimageinpainting}, a contextual reconstruction model. This aids in the reconstruction of lanes that are obstructed due to the presence of vehicular occlusions. 
    
    \item \textbf{Segmentation Node:} Finally, the YOLOPv2 model \cite{han2022yolopv2betterfasterstronger}, a panoptic driving perception system, which segments the lane regions present in the inpainted frame. 
\end{itemize}

% \begin{figure}[h]
%     \centering
%     \begin{tikzpicture}
%         \draw[black, thick] (0,0) rectangle (8,6); % Draws a black rectangle with no fill
%     \end{tikzpicture}
%     \caption{\textcolor{red}{Insert overall network architecture of proposed method here}}
%     \label{fig:black_rectangle}
% \end{figure}

% \noindent We now provide a detailed explanation and implementation details of the nodes mentioned above.
\subsubsection{Detector Node}

% Introduce the core model briefly (here YOLOv8sn)
% Your YOLO implementation detail
% Justification of selecting this particular model

Speed was the primary factor influencing the selection of YOLOv8 model as the detector node, because the main focus of this node was to identify the occlusions that occupy substantial portions of the road area, hence, having lesser emphasis on accuracy. This architecture offers reduced computational overhead and latency, making it particularly suitable for real-time applications. It has 3.01M parameters.

 A pre-trained model of YOLOv8 was fine-tuned using the BDD100k dataset \cite{bdd100k}. The fine-tuning process involved training the model for 20 epochs with a batch size of 23 and a learning rate of 0.000714 using the AdamW optimizer. The loss function contains multiple components to deal with each component of object detection. The classification branch utilizes Binary Cross-Entropy (BCE) Loss to measure the error in the classification of the detected object. The regression branch, responsible for predicting the bounding box, uses a combination of Objectness Loss (BCE Loss) and Location Loss (Complete Intersection over Union Loss). Objectness Loss calculates the error in detecting whether an object is present in a particular grid cell or not. CIoU loss considers both the overlap and aspect ratio differences between the predicted and ground truth boxes. 
 
To ensure that the pipeline was well-adapted to vehicular occlusions and traffic scenarios, the training focused on 7 classes relevant to vehicle traffic - car, pedestrian, truck, bus, train, motorcycle, bicycle. Fig. \ref{fig:results_pipeline_bdd} illustrates the output generated by the YOLOv8 model, on passing the raw image, which is used for the subsequent conversion into a masked image. This process delineates the regions for the Inpainting node.

\subsubsection{Inpainting Node}
Given the masks of the regions with occlusions, the desired output from this node is reconstructed lanes. These masked images were fed to the model to create an appropriate representation of the lanes. CR-Fill, a state-of-the-art inpainting model, was selected considering its lightweight architecture, fast inference capabilities, and support for fine-tuning. It has 4.05M parameters. 

The model consists of a two-stage end (coarse-to-fine) generator network and a multi-scale deepfill discriminator network. The generator is an attention-free network with an auxiliary task focused on contextual reconstruction. This approach encourages the generator to produce plausible outputs that maintain coherence when reconstructed. The auxiliary branch termed contextual reconstruction (CR) loss, optimizes query-reference feature similarity and a reference-based reconstructor alongside the inpainting generator. The coarse network utilizes an L1 loss and the refinement network is trained using a combination of L1 loss, adversarial loss, and CR loss.

The training utilizes the default starting learning rate of 0.0002 for the Adam optimizer with beta1 and beta2 at 0.5 and 0.999, respectively. The model was trained using the Places: A 10 million Image dataset \cite{zhou2017places} to set the initial weights for further fine-tuning. The model was fine-tuned in the context of lane and road reconstruction with a custom dataset built by filtering frames from CULanes and BDD apart from those already used for training other nodes or validation. The images were cropped to focus on the road regions. Apart from the hyperparameters set for the training, the weight for the L1 loss was set to 5 with the VGG feature matching loss and the discriminator feature matching loss ignored. Additionally, the discriminator was set to iterate twice per generator. The model was fine-tuned for 450 epochs.

\subsubsection{Segmentation}
For this node, the YOLOPv2 model \cite{han2022yolopv2betterfasterstronger} was employed. YOLOPv2 is a state-of-the-art panoptic driving perception system designed to perform traffic object detection, drivable road area segmentation, and lane detection simultaneously. For the purpose of this study, only the lane detection head was utilized. 

The architecture encompasses a multi-branch network design comprising of 38.9M parameters. It employs an encoder-decoder configuration, wherein the E-LAN shared encoder extracts diverse, multi-scale semantic features. These features are fused using a Spatial Pyramid Pooling and Feature Pyramid Network module and then processed for information with localized and precise details using a Path Aggregation Network. 

The loss function is a hybrid loss of dice and focal loss, focusing on hard-to-detect instances and handling class imbalance parallelly. The model's learning rate was initially set to 0.01 with a cosine annealing policy and trained for 300 epochs.

YOLOPv2 was chosen for having state-of-the-art performance in terms of accuracy, speed, and ability to demonstrate strong performance in diverse settings like day and night scenarios. The model's efficient network design and memory allocation allows it to maintain high inference speeds, thus making it highly suitable for real-time lane segmentation.

\begin{table*}[ht!]
  \centering
{\renewcommand{\arraystretch}{1.1}%
  \caption{Performance Comparison of Lane Detection Models on CULanes and BDD100K Datasets}
  \label{tab:perf_comp_combined}
  \begin{tabular}{|c|c|c|c|c|c|c|c|c|}
    \hline
    \multirow{2}{*}{\textbf{Model}} & \multicolumn{4}{c|}{\textbf{CULanes}} & \multicolumn{4}{c|}{\textbf{BDD100K}} \\
    \cline{2-9}
     & \textbf{IOU} & \textbf{Precision} & \textbf{Recall} & \textbf{Dice} & \textbf{IOU} & \textbf{Precision} & \textbf{Recall} & \textbf{Dice} \\
    \hline
    Ultra Fast Lane Detection & 0.054 & 0.175 & 0.070 & 0.098 & 0.106 & 0.267 & 0.144 & 0.180 \\
    \hline                  
    LaneNet & 0.077 & 0.272 & 0.105 & 0.137 & 0.156 & 0.332 & 0.232 & 0.255 \\
    \hline
    YOLOPv2 & 0.246 & 0.579 & 0.291 & 0.375 & 0.442 & \textbf{0.856} & 0.473 & 0.602 \\
    \hline
    aug-Segment & 0.273 & 0.562 & 0.343 & 0.416 & 0.134 & 0.161 & 0.359 & 0.209 \\
    \hline
    LOID & \textbf{0.302} & \textbf{0.595} & \textbf{0.372} & \textbf{0.445} & \textbf{0.533} & 0.819 & \textbf{0.602} & \textbf{0.686} \\
    \hline
  \end{tabular}}
\end{table*}

\section{Datasets \& Evaluation}
\label{sec:data_eval}
\begin{figure} [t]
    \centering
    \begin{tabular}{cc}
        \begin{subfigure}
            \centering
            \includegraphics[width = 0.45\linewidth]{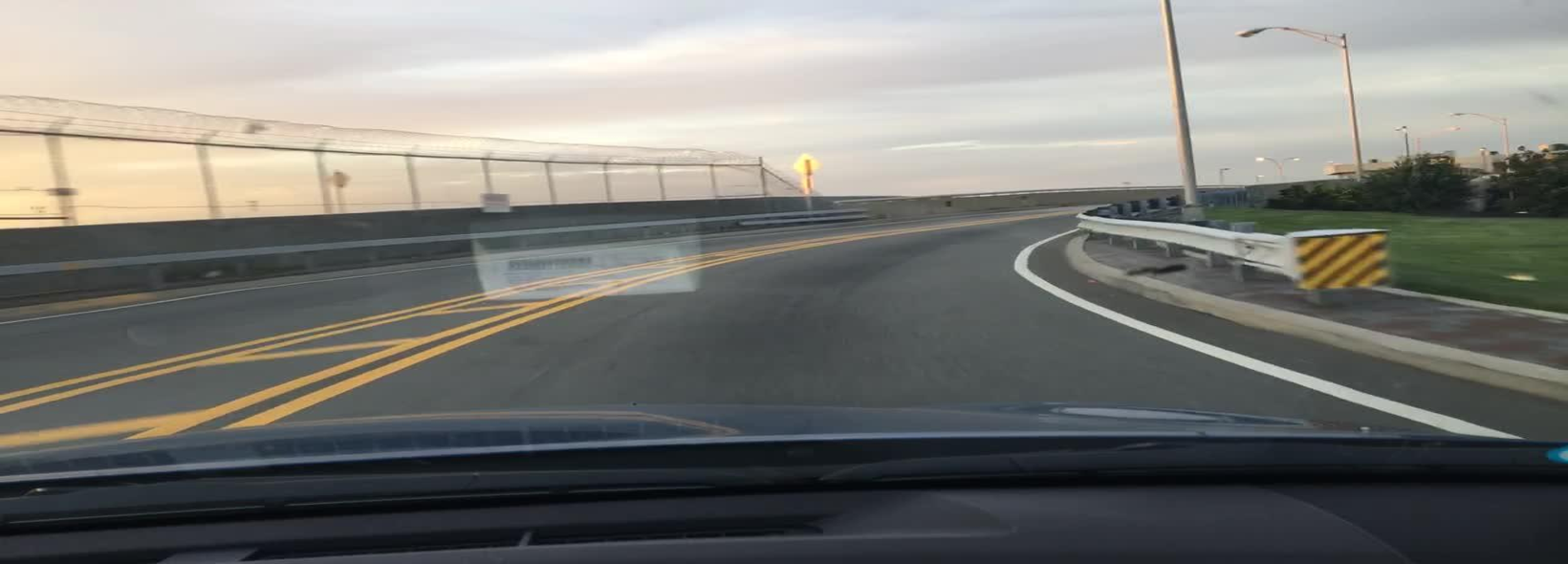}
        \end{subfigure} &
        \begin{subfigure}
            \centering
            \includegraphics[width=0.45\linewidth]{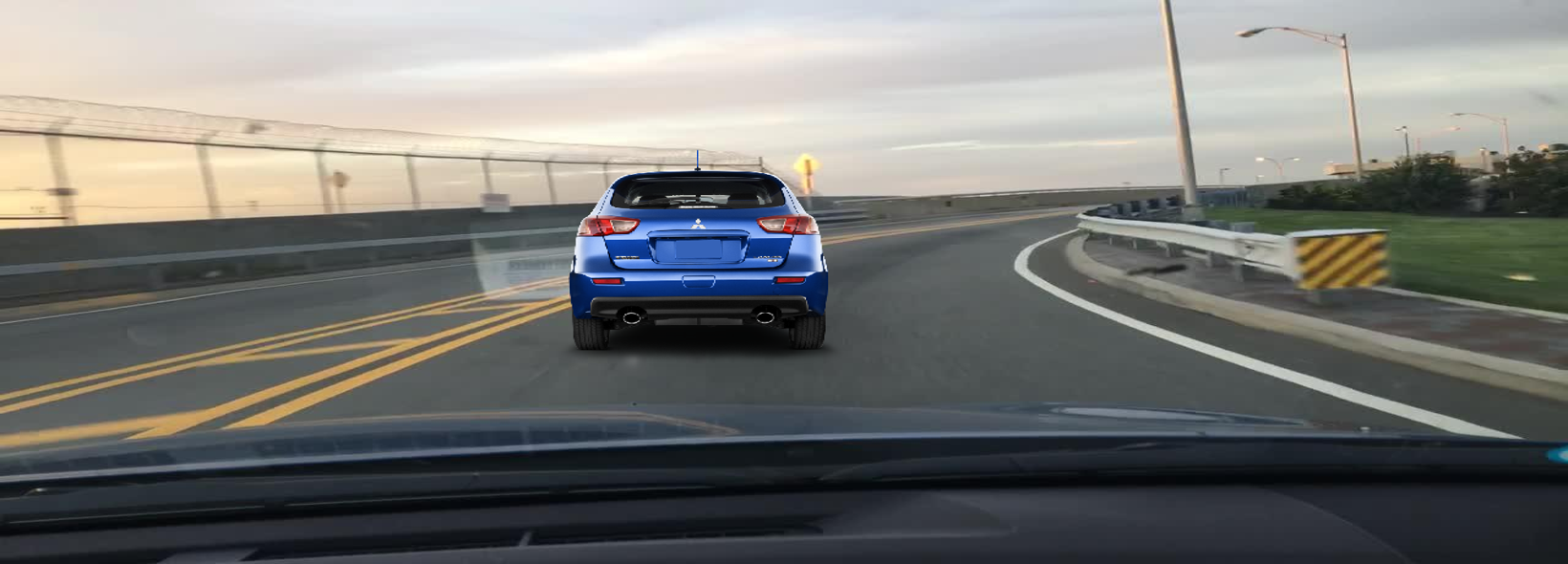} 
        \end{subfigure} \\
        (a) Clear & (b) Augmented Occluded \\ Frame & Frame \\
        
        \begin{subfigure}
            \centering
            \includegraphics[width=0.45\linewidth]{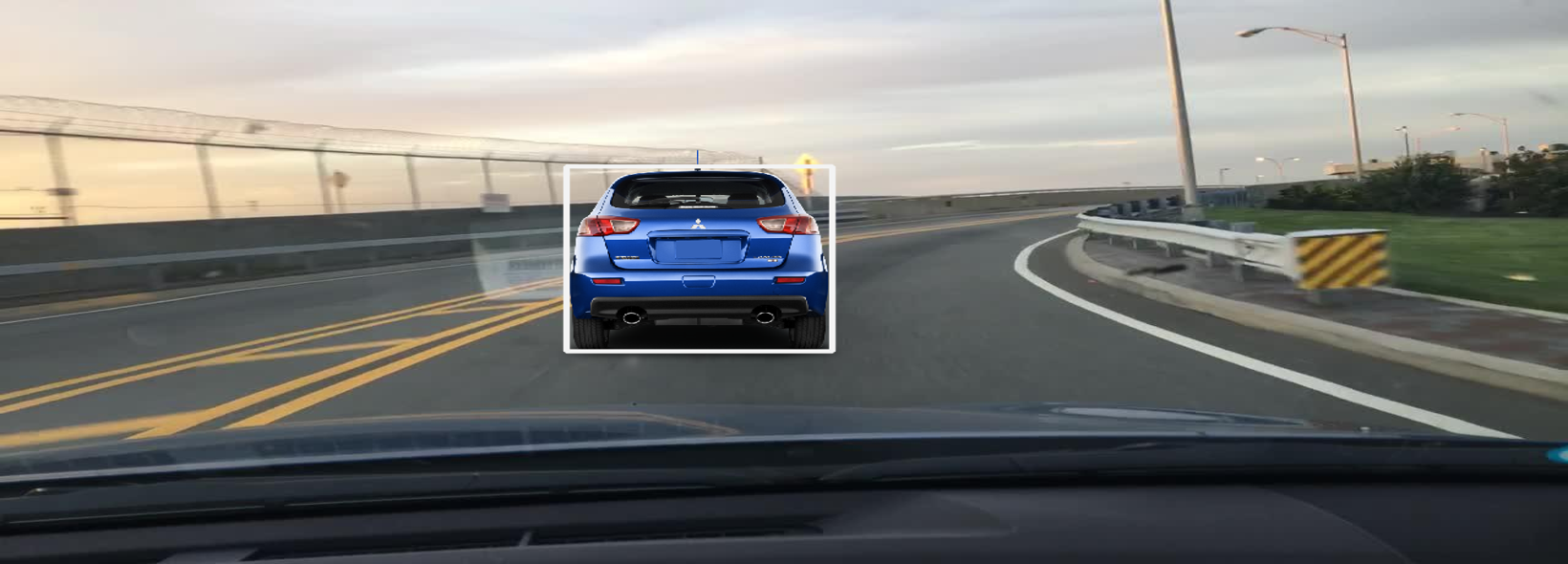}
        \end{subfigure} &
        \begin{subfigure}
            \centering
            \includegraphics[width=0.45\linewidth]{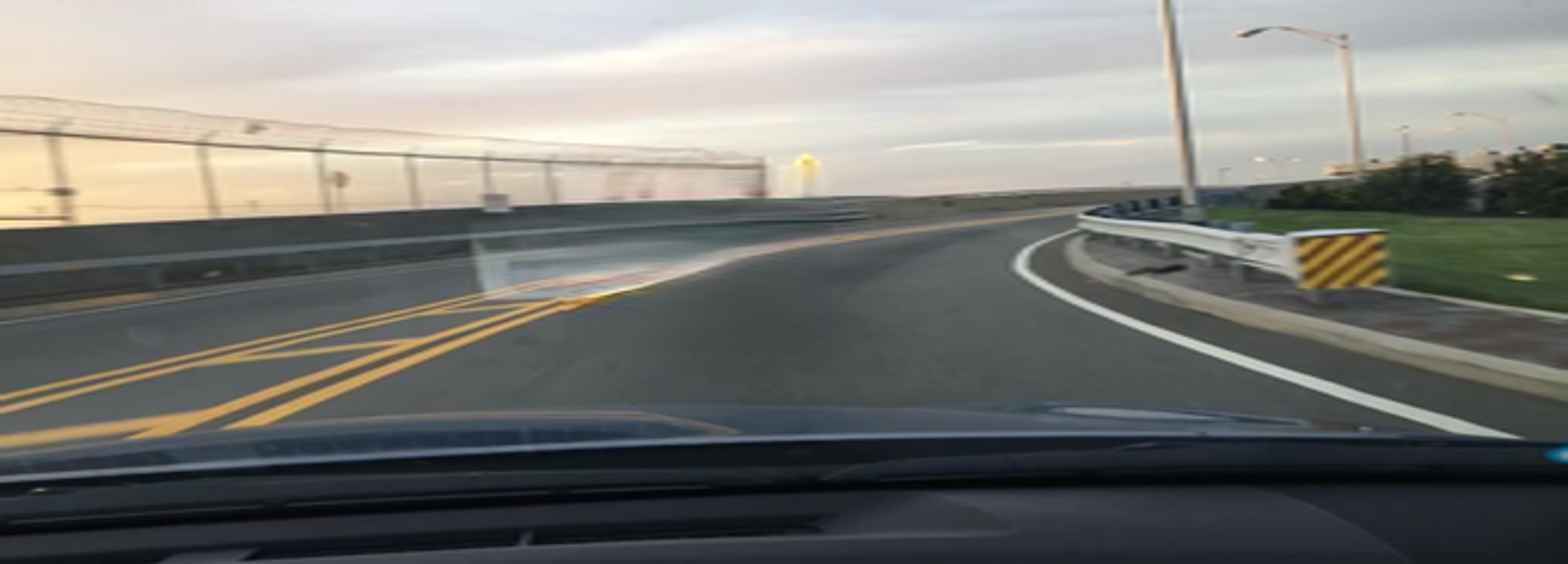}
        \end{subfigure} \\
        (c) Detected Occlusion & (d) Inpainted Frame \\
        
        \begin{subfigure}
            \centering
            \includegraphics[width=0.45\linewidth]{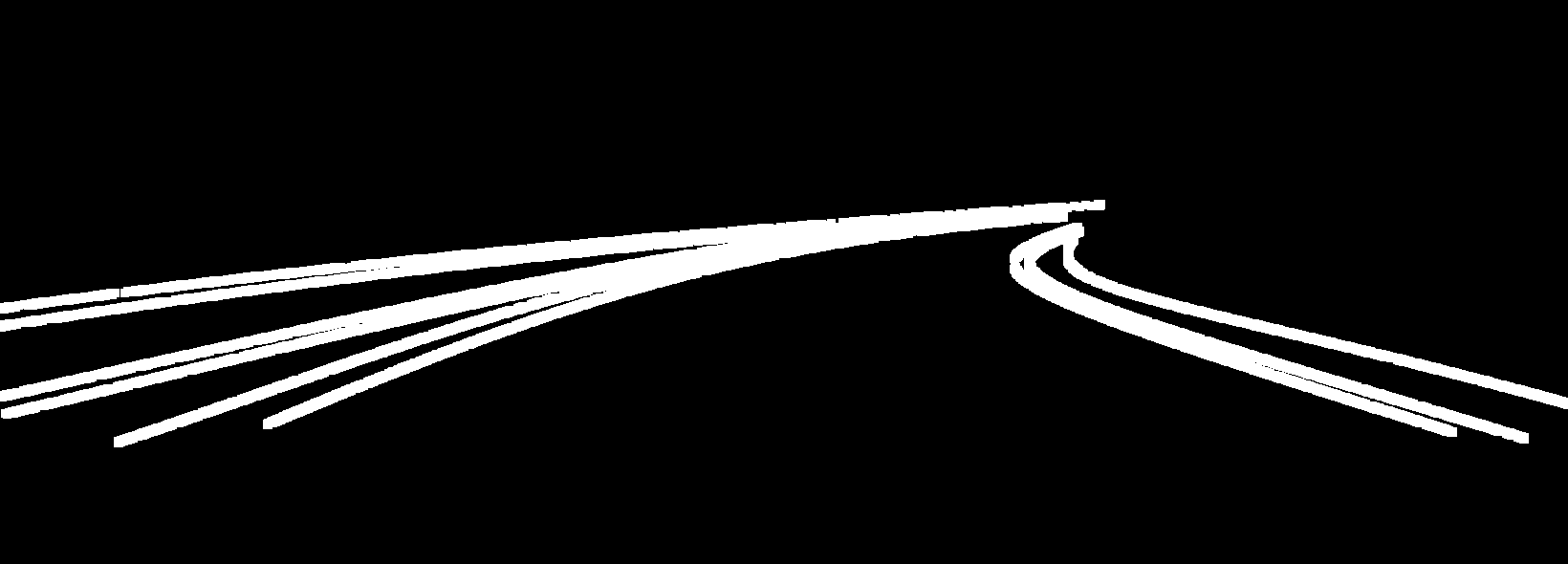}
        \end{subfigure} &
        \begin{subfigure}
            \centering
            \includegraphics[width=0.45\linewidth]{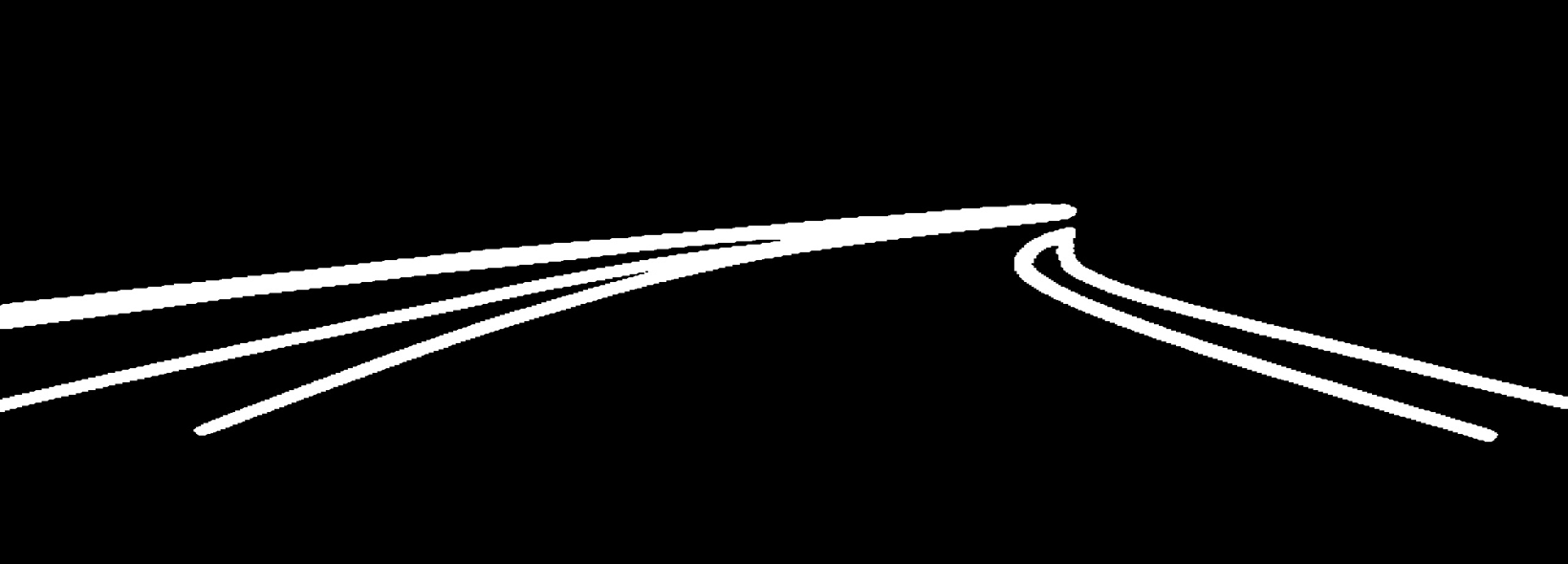}
        \end{subfigure} \\
        (e) Ground Truth  & (f) Clear Frame \\Mask  & Prediciton \\

        \begin{subfigure}
            \centering
            \includegraphics[width=0.45\linewidth]{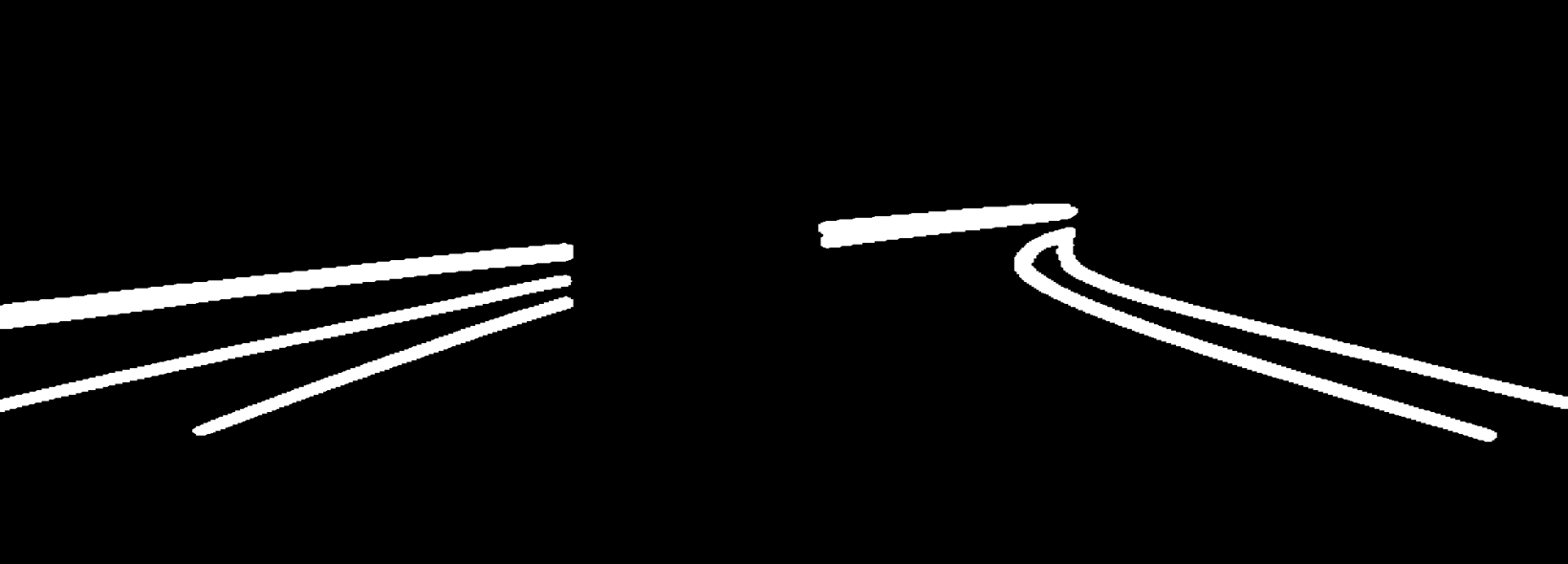}
        \end{subfigure} &
        \begin{subfigure}
            \centering
            \includegraphics[width=0.45\linewidth]{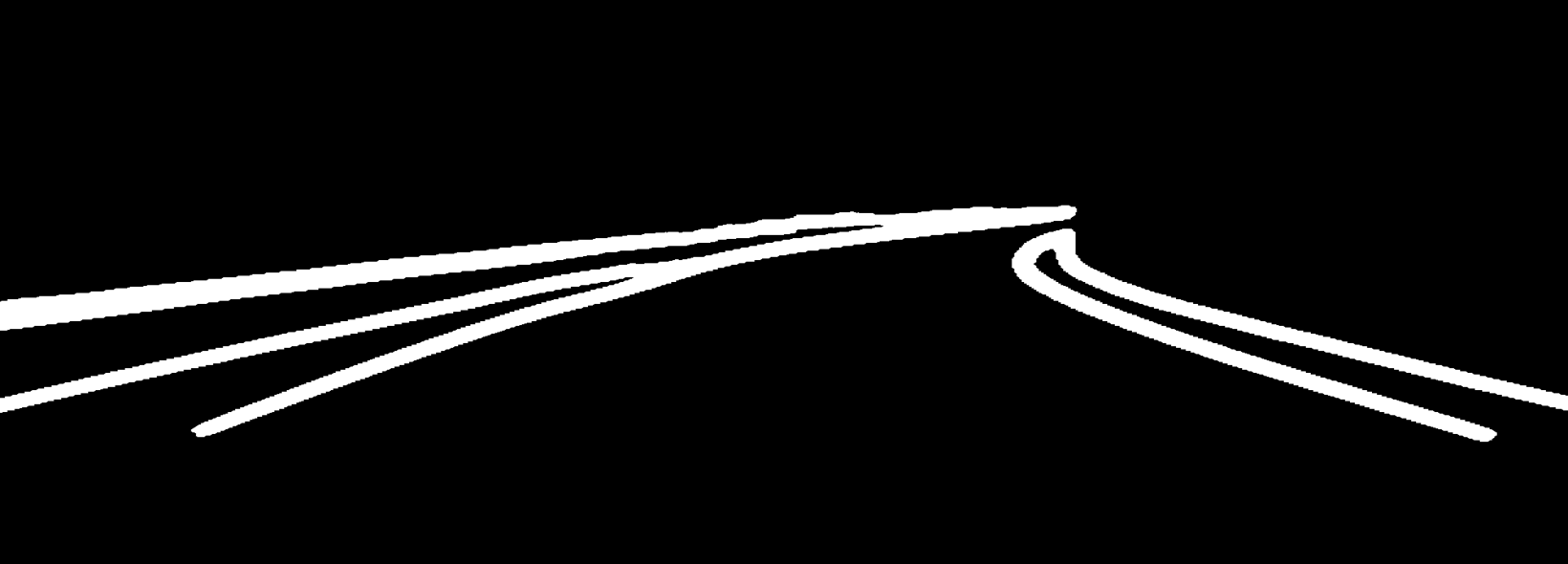}
        \end{subfigure} \\
        (g) Occluded Frame  & (h) Inpainted Frame \\ Prediction & Prediction   \\
    \end{tabular}
    \caption{Pipeline Outputs - BDD}
    \label{fig:results_pipeline_bdd}
\end{figure}

\begin{figure} [t]
    \centering
    \begin{tabular}{cc}
        \begin{subfigure}
            \centering
            \includegraphics[width = 0.45\linewidth]{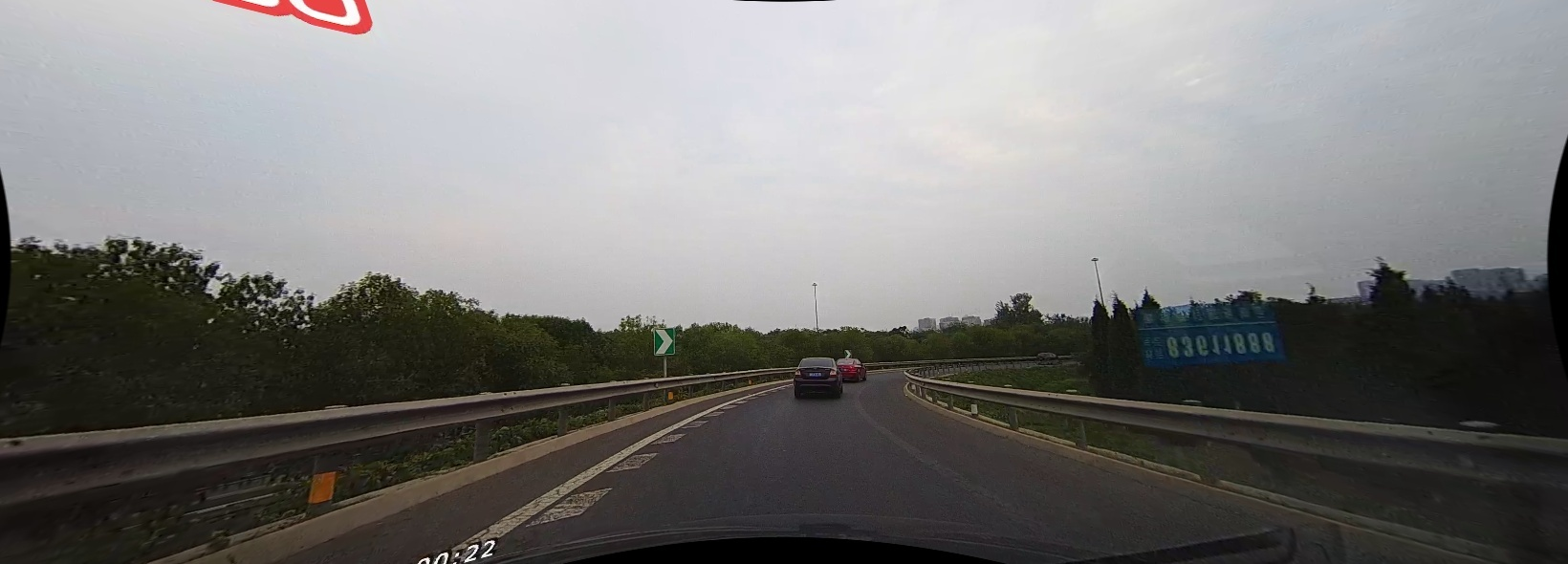}
            % \caption{Caption 1}
        \end{subfigure} &
        \begin{subfigure}
            \centering
            \includegraphics[width=0.45\linewidth]{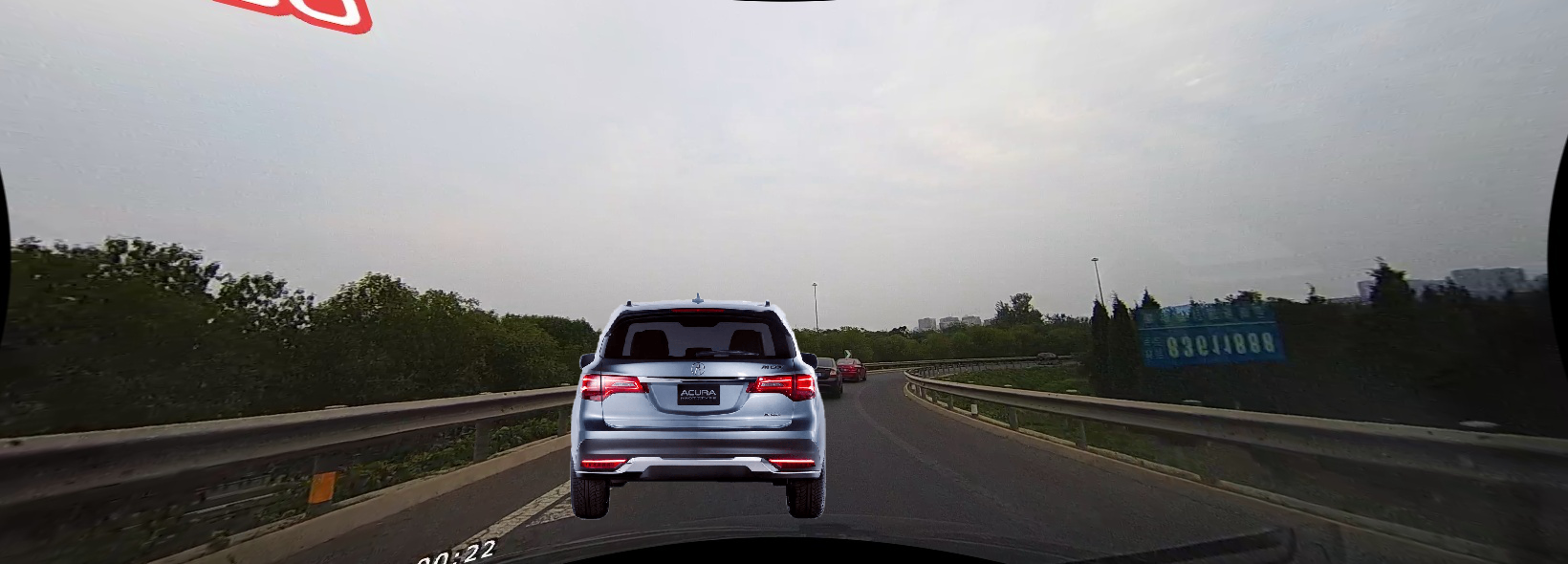} 
            % \caption{Caption 2}
            
        \end{subfigure} \\
        (a) Clear & (b) Augmented Occluded \\ Frame & Frame \\
        \begin{subfigure}
            \centering
            \includegraphics[width=0.45\linewidth]{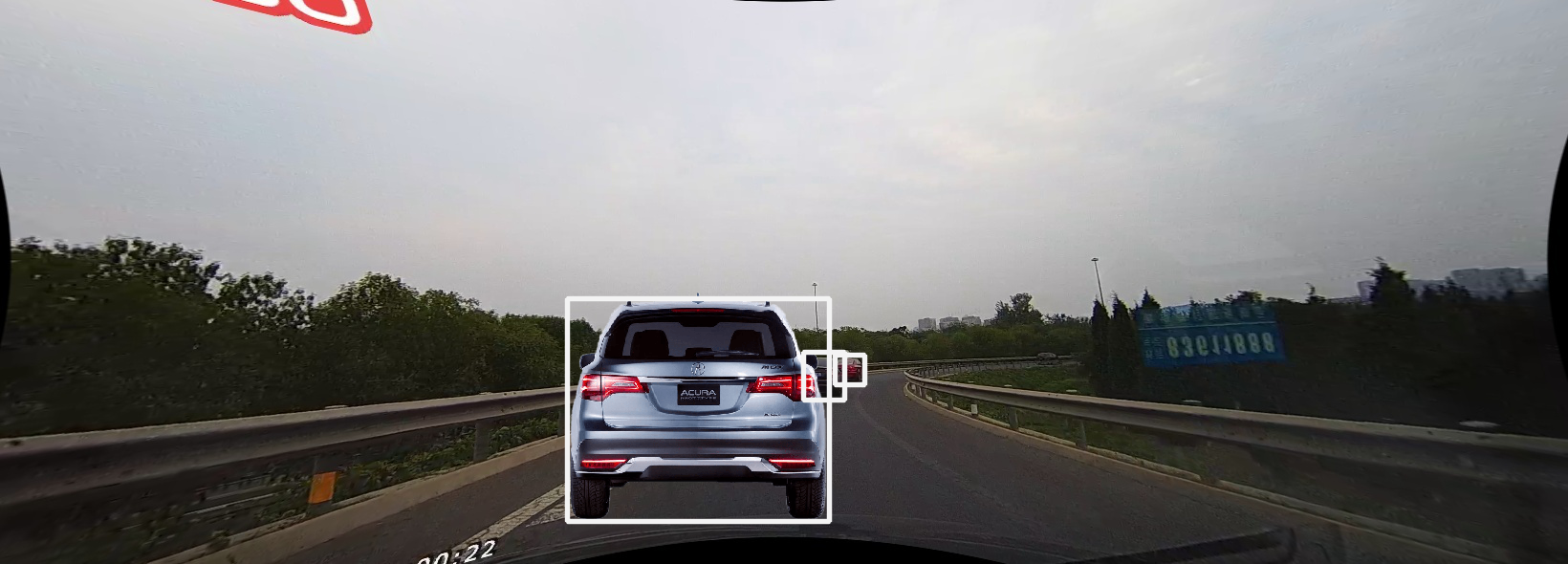}
            % \caption{Caption 3}
        \end{subfigure} &
        \begin{subfigure}
            \centering
            \includegraphics[width=0.45\linewidth]{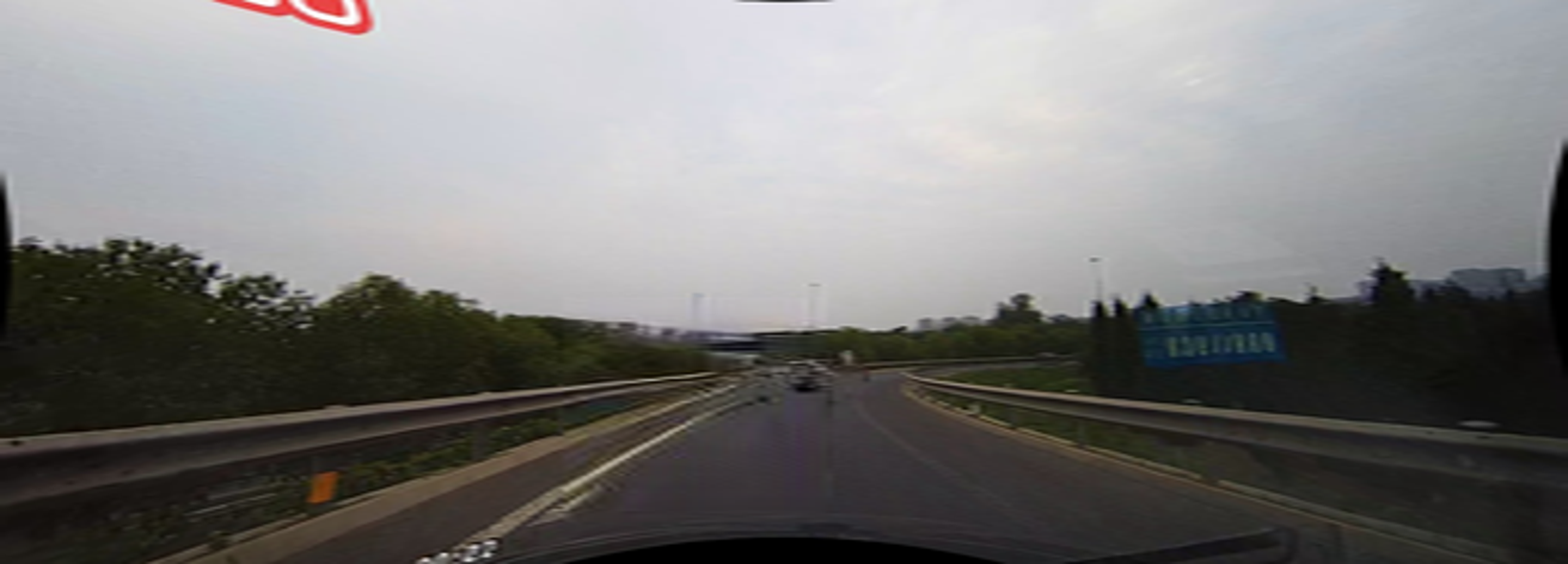}
            % \caption{Caption 4}
            
        \end{subfigure} \\
        (c) Detected Occlusion & (d) Inpainted Frame \\
        \begin{subfigure}
            \centering
            \includegraphics[width=0.45\linewidth]{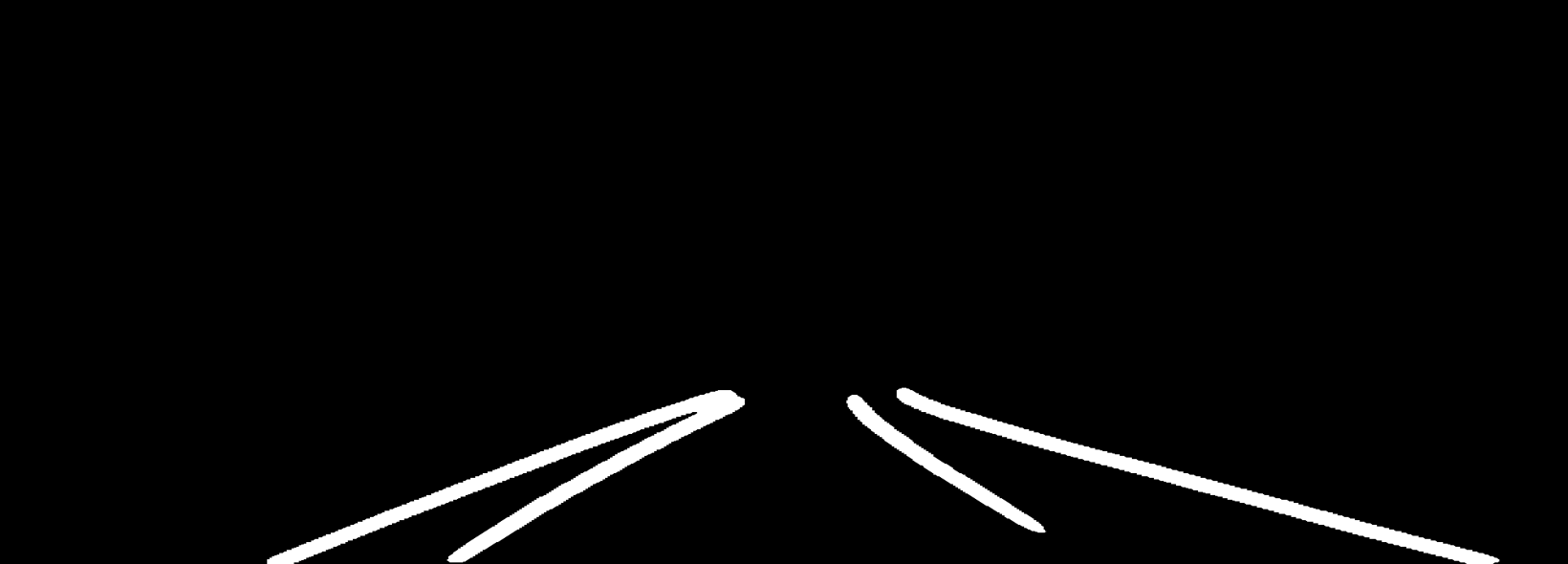}
            % \caption{Caption 5}
        \end{subfigure} &
        \begin{subfigure}
            \centering
            \includegraphics[width=0.45\linewidth]{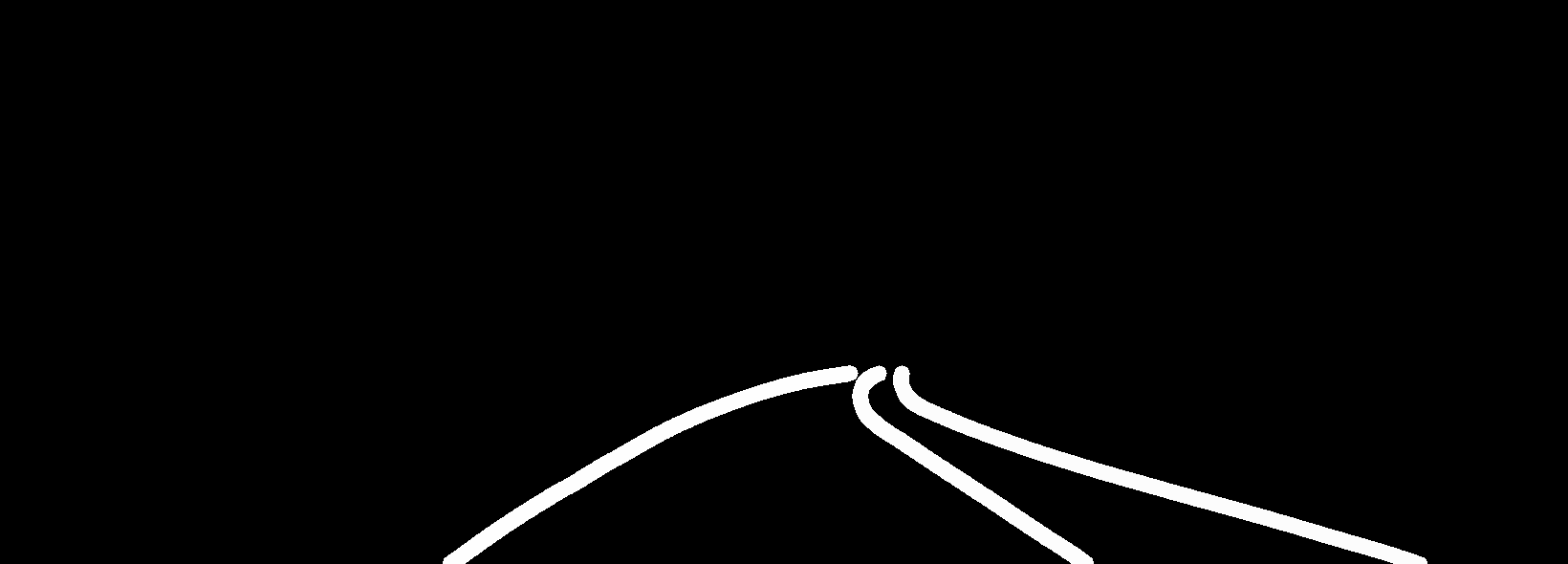}
            % \caption{Caption 6}
        \end{subfigure} \\
        (e) Ground Truth  & (f) Clear Frame \\Mask  & Prediciton \\

        \begin{subfigure}
            \centering
            \includegraphics[width=0.45\linewidth]
            {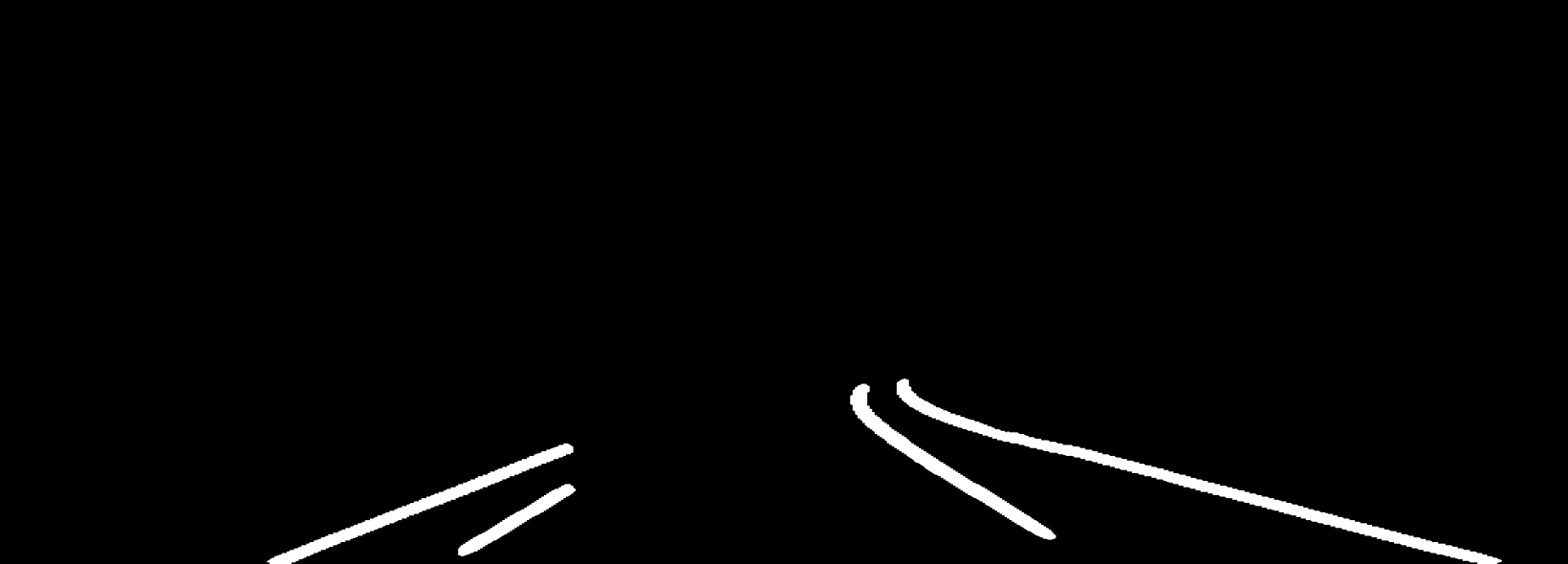}
            % \caption{Caption 7}
        \end{subfigure} &
        \begin{subfigure}
            \centering
            \includegraphics[width=0.45\linewidth]{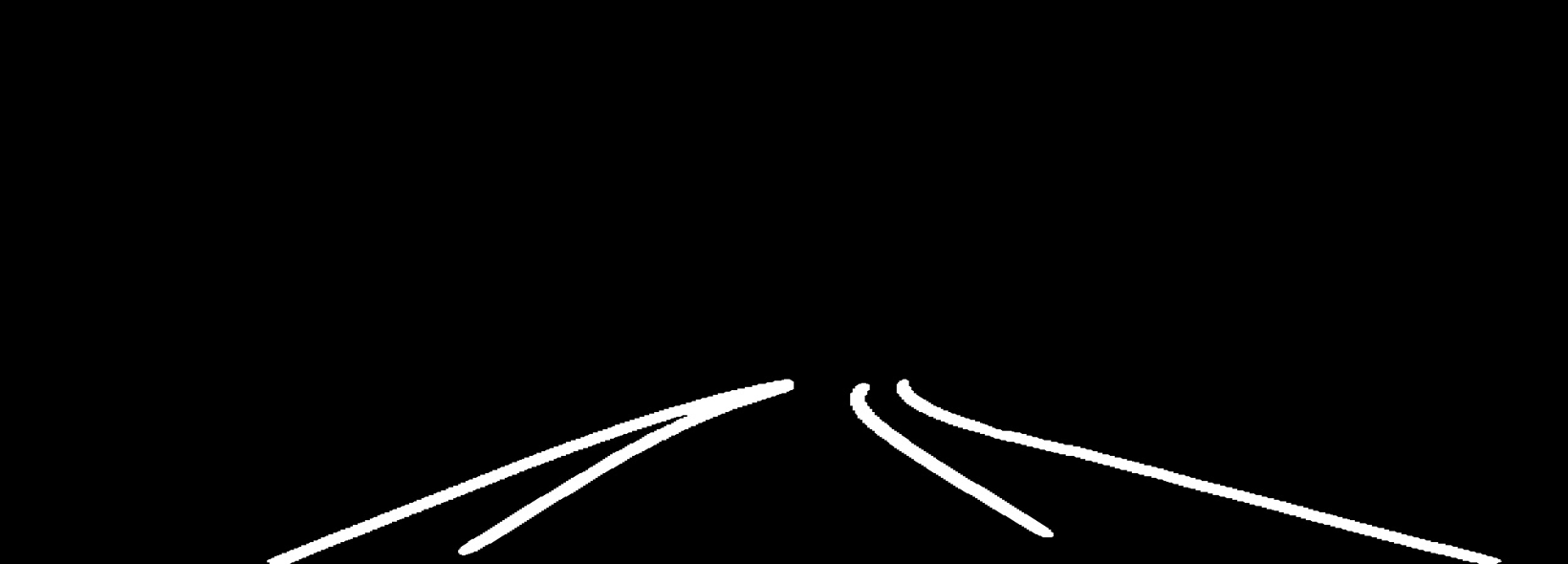}
            % \caption{Caption 8}
        \end{subfigure} \\
        (g) Occluded Frame  & (h) Inpainted Frame \\ Prediction & Prediction   \\
        % Pipeline Results CULanes \\
    \end{tabular}
    \caption{Pipeline Outputs - CULanes}
    \label{fig:results_pipeline_CULanes}
\end{figure}

For the purpose of the study, a custom dataset was built by artificially augmenting occlusions into frames with clearly visible lanes and no occlusions. Firstly, frames (see Fig. \ref{fig:results_pipeline_CULanes}(a) and Fig. \ref{fig:results_pipeline_bdd}(a)) were selected from the CULane \cite{pan2018SCNN} and BDD100k \cite{bdd100k} datasets. The selection criteria ensured that frames with minimal or no occlusions were retained, which ensured complete lane representations in the ground truth masks from the original datasets. These filtered images were then artificially augmented with vehicles on the road. (Fig. \ref{fig:results_pipeline_CULanes}(b) and Fig. \ref{fig:results_pipeline_bdd}(b)). 

\subsubsection{Augmented CULanes Training Set}
\label{cul_train}
The training set, of 1000 images, for the aug-Segment model was built from the CULanes data. 

\subsubsection{Augmented Validation Set}
A custom validation set was created to evaluate the various experiments defined in this study from the CULane \cite{pan2018SCNN} and BDD100k \cite{bdd100k} datasets, yielding 420 and 380 images, respectively. Using this dataset, further validation of the various segments of the study is presented in Section \ref{sec:experiments}.

\subsection{Evaluation}
 The metrics used to evaluate the study are discussed below and evaluated at the pixel level. Precision measures the accuracy of positive detections by comparing the true positive detections against all detected positives. Recall, measures the algorithm's ability to identify actual positives by comparing true positive detections against true positives and false negatives. The Dice coefficient measures the similarity between the predicted and actual regions by comparing the overlap area to the total number of pixels in both regions. IoU further measures this overlap by calculating the ratio of the overlap area to the combined area of the predicted and ground truth regions. 

% \textcolor{red}{ Define TP, FP and FN}\\

% \noindent \paragraph{Precision} \noindent Ratio of true positive detections to the total number of positive detections.
% \begin{equation}
% \text{Precision} = \frac{TP}{TP + FP}
% \end{equation}

% \noindent \paragraph{Recall} Ratio of true positive detections to the total number of actual positives.
% \begin{equation}
% \text{Recall} = \frac{TP}{TP + FN}
% \end{equation}

% \noindent \paragraph{Dice} Ratio of area of overlap divided by the total number of pixels in both predicted and ground truth regions
% \begin{equation}
% \text{Dice} = \frac{2TP}{2TP + FP + FN}
% \end{equation}

% \noindent \paragraph{IoU} Ratio of area of overlap by area of union.
% \begin{equation}
% \text{IoU} = \frac{TP}{TP + FP + FN}
% \end{equation} 

\section{Experiments and Results}
\label{sec:experiments}

\subsection{Experiments}
This section lists the experiments utilized to evaluate the performance of LOID and aug-Segment against other state-of-the-art lane detection models, particularly focusing on roads with and without occlusion. The SOTA models used for evaluation are Ultra Fast Lane Detection \cite{qin2020ultrafaststructureawaredeep}, LaneNet \cite{wang2018lanenetrealtimelanedetection} and YOLOPv2 \cite{han2022yolopv2betterfasterstronger}. All evaluation and inference experiments were performed on an NVIDIA RTX 3060 GPU \& training was performed on an NVIDIA A6000 GPU.

\subsection{Results}
\subsubsection{Quantitative Results}
Table \ref{tab:perf_comp_combined} shows the performance of LOID with respect to other SOTA models on the augmented BDD100K and CULanes Datasets respectively. The inference times are in Table \ref{tab:config_comp}. 
% \textcolor{red}{Bring the mentioned tables to this page}
\begin{table}[H]
  \centering
{\renewcommand{\arraystretch}{1.1}% 
  \caption{Configuration Comparison of Lane Detection Models}
  \label{tab:config_comp}
  \begin{tabular}{|c|c|c|}
    \hline
     \textbf{Model} & \textbf{\# Params} & \textbf{Inference Time}\\
    \hline                  
    Ultra Fast Lane Detection & 6M & 0.004\\
    \hline                  
    LaneNet & 31M &  0.017 \\
    \hline                  
    YOLOPv2 & 39M & 0.035\\
    \hline
    YOLOv8s (Trained) & 7M & 0.021 \\
    \hline
    LOID & 46M & 0.04 \\
    \hline 
  \end{tabular}}
\end{table}

We see all models perform better on the augmented BDDK100 dataset compared to CULanes which can be accredited to the quality of images and variation in annotation style. The annotations for BDD had to be dilated since the dataset only contains lane edge markings. 

The results obtained from aug-Segment indicate a notable performance enhancement on the augmented CULanes dataset, with an approximate improvement of 12\% in the IoU. However, there is a significant 70\% reduction in IoU when evaluated on the augmented BDDK100 dataset. This discrepancy can be attributed to the fact that aug-Segment was specifically trained on a training set created using the CULanes dataset. Consequently, while the model excels in processing frames from the same dataset with similar lane characteristics, it demonstrates limited generalizability across different datasets. Despite this limitation, aug-Segment outperforms the YOLOPv2 model when applied to a smaller, localized dataset. Applications for this can be found in \cite{kim2017multi}.

LOID achieves state-of-the-art performance on the proposed custom datasets. On the augmented CULanes dataset, LOID attains an IOU of 0.302, while on the augmented BDD100K dataset, it reaches an impressive IOU of 0.533. These results demonstrate a significant improvement of approximately 22\% over the next best-performing model, YOLOPv2. Hence, showing the importance of the inpainting node of the model in regaining the lane representations lost due to occlusions.

Interestingly, despite LOID's superior IOU, YOLOPv2 exhibits comparable precision scores. On the BDD100K dataset, YOLOPv2's precision score is even higher than LOID's. This observation suggests that while YOLOPv2 misses out on a lot of pixels from occluded areas, the detection in the clear areas allows for similar or slightly better detection. The inpainting may cause a few false detections in the reconstructed areas. But to get a more comprehensive understanding of the lane representations a higher precision is not enough. 

The dice score for LOID exceeds YOLOPv2 by 15\% \& Lanenet by 190\%. The substantial improvement of LOID over Lanenet and Ultra Fast Lane detection suggests that traditional methods struggle to achieve similar levels of accuracy in real-time lane detection tasks.

While aug-segment provides a reduced inference time, it has a slight performance bump but still is a promising approach. Nonetheless, considering the superior performance of the LOID model in capturing comprehensive lane representations, LOID emerges as the optimal choice for tasks requiring full lane detection capabilities.

\begin{figure*}[t]
    \centering
    \includegraphics[width = \linewidth]{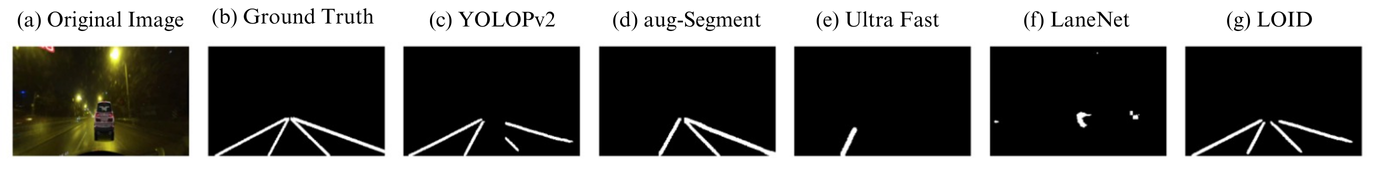}
    \caption{Lane Segmented Outputs from various models | BDD}
    \label{fig:bdd_model_comparision}
\end{figure*}

\begin{figure*}[t]
    \centering
    \includegraphics[width = \linewidth]{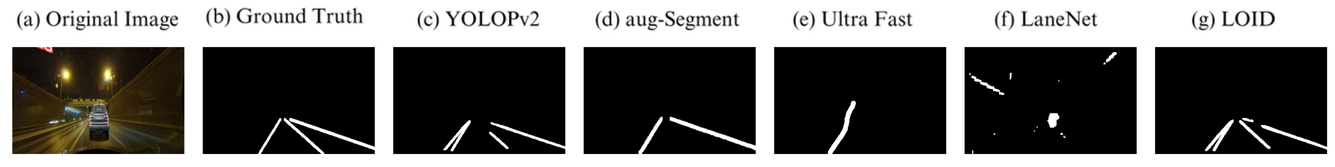}
    \caption{Lane Segmented Outputs from various models | CULanes}
    \label{fig:cu_model_comparision}
\end{figure*}

\subsubsection{Qualitative Results}

Figures \ref{fig:bdd_model_comparision} and \ref{fig:cu_model_comparision} present the results of various models on a sample frame. A noticeable performance improvement is observed with the LOID model, which generates reconstructions that are closest to the ground truth. The LaneNet model struggles with detecting lane-like structures across the image, often being misled by similar patterns. Ultranet and aug-Segment are capable of detecting some occluded lanes and marking them completely; however, they fail to capture all the lanes present in the frame. YOLOPv2 shows good performance on visible data but detects incomplete lanes in occluded regions, leaving gaps where the lanes are not fully visible. In contrast, LOID effectively reconstructs the complete lane representation, including occluded sections, due to its inpainting capabilities. 

The presented figures \ref{fig:results_pipeline_bdd} \& \ref{fig:results_pipeline_CULanes} for augmented BDDK100 and CULanes  show a clear improvement in the results upon using the segmentation model on the detected and inpainted data. The inpainting allows the reconstruction of the missing lane representations. While the lane reconstruction wasn't very detailed and accurate, the detection models trained has the ability to bridge the gap between the inpainted image and the final ground truth mask. 

\subsection{Ablation Study}
For the ablation study, several experiments were conducted to analyze the performance of LOID and evaluate the reliance of each node on the final results.

\begin{table}[H]
  \centering
{\renewcommand{\arraystretch}{1.1}%
  \caption{Performance Metrics for Occlusion Detection}
  \label{tab:object_detection_metrics}
  \begin{tabular}{|c|c|c|c|}
    \hline
    \textbf{Dataset} & \textbf{Precision} & \textbf{Recall} & \textbf{mAP50-95} \\
    \hline                  
    CULanes & 0.868 & 0.955 & 0.592 \\
    \hline
    BDD100k & 0.983 & 0.984 & 0.694 \\
    \hline
    IG02 Dataset & 0.77 & 0.76 & 0.53 \\
    \hline
  \end{tabular}}
\end{table}

This study analyses and reports the individual performance of each component of the pipeline on the custom dataset in comparison with standard datasets. Table \ref{tab:object_detection_metrics} shows the validation metrics of the detector node on the custom dataset and INRIA Annotations for Graz-02 (IG02) \cite{marszalek2007accurate}. The analysis reveals that the performance of the detector node remains consistent between augmented and original datasets, indicating that using an augmented dataset is unlikely to create a significant difference in the overall model performance.

% % Add final numbers for this
% First, an analysis of the detector node has been presented in Table \ref{tab:object_detection_metrics}. This experiment was conducted to see the viability of using an augmented dataset. 

Table \ref{tab:pipeline_metrics_combined} presents the ablation performed on augmented CULanes and BDDK100. The final segmentation node of the model was tested on 4 sets of data. Firstly, clear frames, to see the performance it can provide without occlusions. Further, the occluded dataset was used to evaluate the performance in the case of no inpainting support. Then inpainted data, with results from ground truth bounding box and detector bounding box, are presented to ablate the impact of the detector on the final result. 
% , Clear (Original frames from the dataset), Occluded Frames (Frames augmented with Vehicles), Inpainted with detector bounding box, and Inpainted with Ground Truth Bounding Box. 

% The final segmentation node of LOID was used on 
The clear frame result shows the best results we can achieve using the current segmentation node. The occluded frame results show how much scope for improvement is there. The difference in IOU between these 2 data points are 0.10 \& 0.16 on CULanes and BDDK100 respectively. Using the inpainting solution increases the IoU by 0.06 and 0.09 respectively.

\begin{table*} 
  \centering
{\renewcommand{\arraystretch}{1.1}%
  \caption{Ablation Study -- Segmentation Node of LOID}
  \label{tab:pipeline_metrics_combined}
  \begin{tabular}{|c|c|c|c|c|c|c|c|c|}
    \hline
    \multirow{2}{*}{\textbf{Data}} & \multicolumn{4}{c|}{\textbf{CULanes}} & \multicolumn{4}{c|}{\textbf{BDD100K}} \\
    \cline{2-9}
     & \textbf{IOU} & \textbf{Precision} & \textbf{Recall} & \textbf{Dice} & \textbf{IOU} & \textbf{Precision} & \textbf{Recall} & \textbf{Dice} \\
    \hline
    Clear (Original) & 0.345 & 0.578 & 0.451 & 0.493 & 0.600 & 0.828 & 0.684 & 0.743 \\
    \hline                  
    Occluded (Augmented) & 0.246 & 0.579 & 0.291 & 0.375 & 0.442 & 0.856 & 0.473 & 0.602 \\
    \hline
    Inpainted with Detector bb. & 0.302 & 0.595 & 0.372 & 0.445 & 0.533 & 0.819 & 0.602 & 0.686 \\
    \hline
    Inpainted with Ground Truth bb. & 0.291 & 0.562 & 0.385 & 0.412 & 0.530 & 0.789 & 0.619 & 0.685 \\
    \hline
  \end{tabular}}
\end{table*}

The precision score presents interesting results a trend where precision is the same or slightly higher for the occluded frames can be noticed. This is because the model has more lane ``area'' to predict which causes it to make mistakes henceforth dropping the precision. A similar trend can be seen with the inpainted data as well. We can therefore infer that the segmentation node has a higher impact on the precision score than the inpainting node. Both forms of data (inpainted and clear) show lower or similar precision scores compared to occluded frames.

% The inpainted results from the detector bound box have an IoU value of about 73\% of the clear frame predictions.

When comparing the Intersection over Union (IoU) of the inpainted data with the bounding boxes generated by the detector and the ground truth, we observe comparable results across both datasets, attributable to the high accuracy of the detector model. Furthermore, slightly dilated predictions—where the bounding box predictions are marginally larger—have contributed to an enhancement in accuracy relative to the ground truth model.

% \ref{tab}

\section{Discussion and Future Work}
\label{sec:discuss}

The proposed ``aug-Segment'' approach improves model training for lane detection task, through enriched training dataset. However it performs well only on it's tuned dataset (CULanes) and transfers poorly to others (e.g., BDD100K). Our second approach, LOID addresses these limitations, offering significant performance gains across both datasets. Although LOID outperforms other models considerably, its higher inference time may limit its use. While adequate for autonomous driving, increasing the frame rate could further enhance performance, especially for applications requiring faster processing.
%The proposed ``aug-Segment'' approach presents a better approach to training models in a more robust way and this approach is something that can be used across other machine learning applications. However, the performance is not ideal and it lacks robustness. Since it performs better on the dataset that it has been tuned to (CULanes) and this does not transfer well to the other dataset (BDDK100). LOID can deal with all these issues and gives a significant performance improvement. While LOID outperforms all the other models by a considerable margin on both datasets, it lacks a high inference time. While the current inference time presented is enough for autonomous driving applications, a higher frame rate can improve performance even better. Some applications might require a higher frame rate. 

Future work can include merging certain aspects of the pipeline such as the detector and inpainting node. Essentially, sending the detected and masked image to the latent space on the model. This might help increase inference time. Additionally, this paper largely deals with vehicular occlusions future work can involve studies on how LOID performs with other occlusions such as pedestrians, potholes, and weather-caused occlusions (rain, fog, and snow).

\section{Conclusion}
\label{sec:conclusion}
This paper proposes two methods to deal with road occlusions during lane detection: aug-Segment and LOID. The ``aug-Segment'', a lightweight model trained by utilizing occlusion-augmented datasets. This approach resulted in an improvement of approximately 12\% in the mIoU compared to SOTA models using similar source data. The model offers a low inference time, particularly well-suited for applications where consistent frame rates are expected but high inference times are a concern.

Next we  present LOID a lane detection pipeline, that deals with conclusions from the outset by detecting and inpainting them and before performing lane detection. LOID is highly adaptable; allowing for the customization and tuning of individual components to meet specific requirements of various applications.
%any nodes can be swapped out or tuned to match the specific requirements better for the particular application. 
Using the detection and inpainting pipeline improved the mIOU for the YOLOPv2 model by 22\% with a marginal increase in the inference time. The inference time for LOID is 0.04 seconds which is 14\% higher than YOLOPv2, LOID can be run at 25 frames per second. Given that a frame rate of 25 frames per second is sufficient for Advanced Driver Assistance Systems (ADAS), as demonstrated in \cite{horgan2015vision}, using LOID is viable for real-time driving applications.

\bibliographystyle{IEEEtran}
\bibliography{biblio}

% \begin{comment}
% \newpage

% If you have an EPS/PDF photo (graphicx package needed), extra braces are needed around the contents of the optional argument to biography to prevent the LaTeX parser from getting confused when it sees the complicated \texttt{includegraphics} command within an optional argument. (You can create your own custom macro containing the \texttt{includegraphics} command to make things simpler here.)


% Use \texttt{begin\{IEEEbiography\}} and then for the 1st argument use \texttt{includegraphics} to declare and link the author photo. Use the author name as the 3rd argument followed by the biography text.
% \end{IEEEbiography}


% Use \texttt{begin\{IEEEbiographynophoto\}} and the author name as the argument followed by the biography text.
% \end{comment}

\end{document}